\documentclass{article}
\usepackage[utf8]{inputenc}
\usepackage{setspace}
\usepackage{diagbox}

\usepackage{multirow}

\usepackage{amsfonts}
\usepackage{amssymb}
\usepackage{mathrsfs}

\usepackage{rotating}
\usepackage{adjustbox}
\usepackage{spconf,amsmath,graphicx}

\usepackage{booktabs}       
\usepackage{amsfonts}       
\usepackage{nicefrac}       
\usepackage{microtype}      
\usepackage{color}
\usepackage{algorithm,algpseudocode}
\usepackage{subcaption}
\usepackage{graphicx}
\usepackage{amsmath}
\usepackage{float}
\usepackage{amssymb}
\usepackage{pifont}

\newcommand{\xmark}{\ding{55}}%
\usepackage{tabularx}
\usepackage{amsmath}
\usepackage{rotating}
\usepackage{adjustbox}
\usepackage{latexsym}
\usepackage{array}
\usepackage{booktabs}
\usepackage{bbding}
\usepackage{lipsum}

\usepackage{makecell}

\title{Bayesian Transformer Language Models for Speech Recognition}

\name{Boyang Xue$^*$, Jianwei Yu$^*$\thanks{Equal contribution}, Junhao Xu, Shansong Liu, Shoukang Hu, Zi Ye, \protect\\ \textit{Mengzhe Geng, Xunying Liu, Helen Meng}}
\address{$^1$The Chinese University of Hong Kong \\
\texttt{\{byxue,jwyu,jhxu,ssliu,skhu,zye,mzgeng,xyliu,hmmeng\}@se.cuhk.edu.hk}}
\begin{document}
\ninept
\maketitle
\begin{abstract}
State-of-the-art neural language models (LMs) represented by Transformers are highly complex. Their use of fixed, deterministic parameter estimates fail to account for model uncertainty and lead to over-fitting and poor generalization when given limited training data. In order to address these issues, this paper proposes a full Bayesian learning framework for Transformer LM estimation. Efficient variational inference based approaches are used to estimate the latent parameter posterior distributions associated with different parts of the Transformer model architecture including multi-head self-attention, feed forward and embedding layers. Statistically significant word error rate (WER) reductions up to 0.5\% absolute (3.18\% relative) and consistent perplexity gains were obtained over the baseline Transformer LMs on state-of-the-art Switchboard corpus trained LF-MMI factored TDNN systems with i-Vector speaker adaptation. Performance improvements were also obtained on a cross domain LM adaptation task requiring porting a Transformer LM trained on the Switchboard and Fisher data to a low-resource DementiaBank elderly speech corpus. 

\end{abstract}
\section{Introduction}
\label{sec:intro}

Language models (LMs) play an important role in automatic speech recognition (ASR) systems and many other applications. Language models compute the joint probability of a given sentence ${\bf W}=({{\bf w}_1,{\bf w}_2,...,{\bf w}_T})$ as:
\vspace{-2.0em}

\begin{equation}
\small
p({\bf W}) = p({{\bf w}_1,{\bf w}_2,...,{\bf w}_n})=\prod_{t=1}^{n} P({{\bf w}_t|{\bf w}_{t-1},..., {\bf w}_1})
\end{equation}
\vspace{-1.4em}

\noindent which can be expressed using the multiplication of word level probabilities. The key part of the statistical language modelling problem is to learn long-range context dependencies. Directly modelling long-span contexts lead to a severe data sparsity problem for $n$-gram language models \cite{telrm}. To this end, neural language models that can represent longer span preceding history contexts in a continuous vector space, for example, based on long-short term memory recurrent neural networks (LSTM-RNNs) \cite{rnnblm,iteernnlm} can be used. 

In recent years deep Transformer models \cite{aiayn} have defined state-of-the-art language modelling performance across a range of speech recognition tasks \cite{lmwdt}. The Transformer model architecture features a deep stacking of multiple self-attention layers \cite{lstmnml,assse,adamnl} with residual connections \cite{drlir} and layer normalization \cite{ln} to learn long-range contexts. Positional encoding layers \cite{aiayn,cssl} are used to further augment the self-attention layers with sequence order information. Performance improvements over the conventional LSTM-RNN language models have been widely reported \cite{lmwdt,aestnlma}. 

The highly complex neural architecture design of Transformers often leads to a large increase in the overall system complexity, for example, up to hundreds of millions of parameters \cite{lmwdt}. In common with other deep learning based language modelling approaches \cite{rnnblm,iteernnlm}, the use of fixed, deterministic parameter estimates in conventional Transformer models fails to account for model uncertainty. When given limited training data, standard Transformer models are prone to over-fitting and poor generalization.  This issue can be further aggregated when rapidly adapting a well-trained Transformer model to small size dataset associated with a new style, genre or domain \cite{aestnlma}. The current solution to this problem is largely based on dropout \cite{daswpnno}, a simple and effective regularization approach used in many deep learning systems including neural network language models \cite{brnnlm,gplrnnlm,csprum}. However, it lacks of a mathematically well-defined framework \cite{dbarmudl,blamnnu}. The underlying dropout distribution also requires hyper-parameter setting on an empirical basis for different tasks. 

In order to address these issues, this paper proposes a full Bayesian learning framework to account for model uncertainty in Transformer language model estimation. An efficient variational inference based approach is adopted to estimate the latent parameter posterior distribution. A systematic investigation on the effects of performing Bayesian estimation in different parts of the Transformer model architecture including the self-attention, feed forward and embedding layers is performed. Statistically significant word error rate (WER) reductions up to 0.5\% absolute (3.18\% relative) were obtained over the baseline Transformer LM on a state-of-the-art 900 hour speed perturbed Switchboard corpus trained LF-MMI factored TDNN system with i-Vector speaker adaptation \cite{lhuc}. Consistent performance improvements were also obtained on a cross domain LM adaptation task requiring rapidly porting a Transformer LM trained on Switchboard and Fisher data to a small size DementiaBank elderly speech corpus. 

The main contributions of this paper are summarized as follows. First, to the best of our knowledge, this paper is the first work to apply Bayesian learning methods to Transformer language models for speech recognition tasks. In contrast, the only previous research on Bayesian Transformer \cite{blamnnu,blt} was conducted on machine translation and probabilistic programming tasks. Prior works on uncertainty modelling under the Bayesian framework for neural network language modelling approaches were limited to RNNs \cite{brnnlm} and their LSTM or GRU based variants \cite{csprum}.

The rest of this paper is organized as follows. Section 2 reviews the conventional Transformer based language models. Section 3 presents Bayesian Transformer language models. Implementation issues are discussed in Section 4. Experiments and results are shown in section 5. Finally, conclusions and future work are discussed in section 6. 

\vspace{-0.2cm}
\section{Transformer Language models}
The original Transformer architecture proposed in \cite{aiayn} for neural machine translation contains an encoder and a decoder.
In this work, following \cite{lmwdt, aestnlma, ilugp}, the decoder component inside the Transformer architecture was adopted for language modelling.

As shown in Figure 1, the Transformer language model used in this work is a stack of 6 Transformer decoder blocks. 
Each block consists of a multi-head self-attention \cite{assse,adamnl} module and a feed forward module.
Residual connections \cite{drlir} and layer normalization \cite{ln} are also inserted between these two modules.
Let ${\bf x}_t^{l-1}$ denotes the output of the $(l-1)$-th Transformer block at time $t$. The multi-head self-attention module in the $l$-th block transforms ${\bf x}_t^{l-1}$ to ${\bf z}_t^{l}$  is given as follows:
\begin{align}
    {\bf q}_{t}^{l}, {\bf k}_{t}^{l}, {\bf v}_{t}^{l}&=\mathbf{Q}{\bf x}_{t}^{l-1},\mathbf{K}{\bf x}_{t}^{l-1},\mathbf{V}{\bf x}_{t}^{l-1} \\
    {\bf h}_{t}^{l} &= ({\bf h}_{t-1}^{l}, ({\bf k}_{t}^{l},{\bf v}_{t}^{l})) \\
    {\bf y}_{t}^{l} &= \mathbf{W_h}^{l}\text{SelfAttention}({\bf h}_{t}^{l}, {\bf q}_{t}^{l}) + {\bf x}_{t}^{l-1} \\
    {\bf z}_{t}^{l} &= \text{LayerNorm}({\bf y}_{t}^{l})
\end{align}
where ${\bf Q}$, ${\bf K}$, ${\bf V}$ are projection matrices which map the input ${\bf x}_t^{l-1}$ into query ${\bf q}_t^l$, key ${\bf k}_t^l$ and value ${\bf v}_t^l$ respectively.
${\bf h}_t^l$ is the sequence of of cached key-value pairs up to time $t$, which only contains the history context information and can prevent the model from using any future context.
$\text{SelfAttention}$ denotes the scaled multi-head dot product self-attention \cite{aiayn}.
$\text{LayerNorm}$ represents the layer normalization operation \cite{ln}.
${\bf W}_{h}$ denotes the projection matrix applied to the outputs of the $\text{SelfAttention}$ operation for residual connection \cite{drlir}. The normalized output ${\bf z}_{t}^{l}$ is then fed into the feed forward module:
\begin{align}
    {\bf s}_{t}^{l} &= \mathbf{W_2}^{l}\text{GELU}(\mathbf{W_1^{l}}{\bf z}_{t}^{l}) + {\bf z}_{t}^{l} \\
    {\bf x}_{t}^{l} &= \text{LayerNorm}({\bf s}_{t}^{l})
\end{align}
In this work, the Gaussian error linear unit (GELU) activation function \cite{bnsrgelu} is adopted as the activation function in the feed forward module.

\begin{figure}[h]
\newcommand{\tabincell}[2]{\begin{tabular}{@{}#1@{}}#2\end{tabular}}
\setlength{\parskip}{0.0ex}

  \centering
  \centerline{\includegraphics[width=9cm]{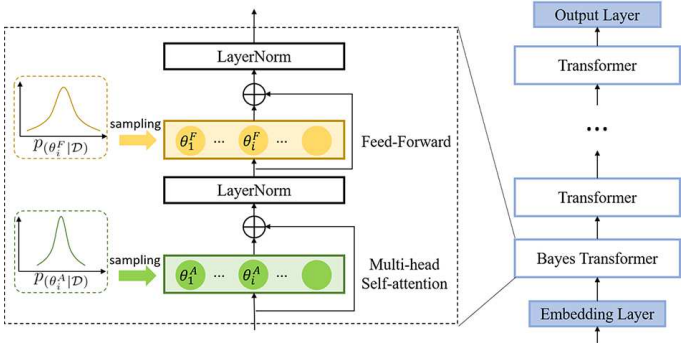}}

\caption{An illustration of the proposed Bayesian Transformer language model architecture. $\theta_{i}^{F}$ and $\theta_{i}^{A}$ denotes the Bayesian model parameters in the feed forward and multi-head self-attention module respectively. }
\label{fig:res}
\vspace{-1.8em}
\end{figure}

\section{Bayesian Transformer LMs}
\label{sec:BTMLM}
In this section, we first propose the formulation of the Bayesian Transformer LM and then present an efficient training scheme based on vairational inference \cite{gpnnsr,hubayesian} for the proposed model. 
\subsection{Bayesian Neural Language Model}
Although Transformer LMs have demonstrated state-of-the-art performance on many speech recognition tasks, the use of fixed-point parameter estimates in these models fails to account for the model uncertainty associated with the words prediction.
When given limited training data, standard Transformer models are prone to over-fitting and poor generalization.
To model the parameter uncertainty in Transformer LMs, Bayesian neural networks can be adopted to treat the model parameters ${\bf \Theta}$ as a posterior probability distribution $p({\bf \Theta}|\mathcal{D})$.
Given the word history context, the word prediction at frame $t$ is computed as follows:
\vspace{-0.5em}
\begin{align}
    p({\bf w}_t|{\bf w}_1, ..., &{\bf w}_{t-1}) \nonumber \\
                                              &=\int p({\bf w}_t|{\bf w}_1, ..., {\bf w}_{t-1}, {\bf \Theta})p({\bf \Theta}|\mathcal{D}) d{\bf \Theta} 
\end{align}
\vspace{-0.5em}

\noindent where $\mathcal{D}$ represents the whole training set for model development and $p({\bf \Theta}|\mathcal{D})$ denotes the posterior distribution of the model parameters learned from the training data.

\vspace{-1.0em}

\subsection{Variational Training for Bayesian Transformer LMs}
To estimate the posterior distribution of the model parameters $p({\bf \Theta}|\mathcal{D})$, the usual approach in Bayesian learning is to maximize the marginal probability.
However, computing this marginal distribution $\mathcal{L}$ is intractable under the Transfomer LM framework. 
Thus, the following variational lower bound is often adopted as an approximation \cite{aevb}:
\vspace{-1.5em}

\begin{small}
\begin{align}
&\log p({\mathcal{D}}) = \log\int p({\bf \mathcal{D}}|{\bf \Theta})p_{r}({\bf \Theta}) d {\bf \Theta} \\
                    &\ge \underbrace{\sum_{n=1}^{N}\log\int p({\bf W}^{n}|{\bf \Theta})q({\bf \Theta}) d {\bf \Theta}}_{\mathcal{L}_{1}} - \underbrace{\text{KL}(q({\bf \Theta})||p_{r}({\bf \Theta}))}_{\mathcal{L}_{2}} = \mathcal{L}
\end{align}
\end{small}
\vspace{-1em}

\noindent where ${\bf W}^{n}$ denotes the $n$-th sentence in the training set $\mathcal{D}$ and $N$ is the total number of sentence in the training set.
$q({\bf \Theta})$ is the variational approximation of the parameter posterior distribution $p({\bf \Theta}|\mathcal{D})$, $p_r({\bf \Theta})$ is the prior distribution of ${\bf \Theta}$ and $\text{KL}(\cdot||\cdot)$ denotes the Kullback-Leiber (KL) divergence.
As shown in Equation (10), the variational lower bound can be decomposed into two parts:
1) the expectation of the log likelihood of the word sequence ${\bf W}$ over the approximated posterior distribution $q({\bf \Theta})$;
2) the KL divergence between $q({\bf \Theta})$ and the prior distribution $p_r({\bf \Theta})$.
Equation (10) is used as the objective function during the model training process. 

As commonly adopted in \cite{blamnnu}, both $q({\bf \Theta})$ and $p_r({\bf \Theta})$ are assumed to be diagonal Gaussian distributions in this work: 
\begin{align}
    q({\bf \Theta})=\mathcal{N}({\bf \Theta}; {\pmb\mu}, {\pmb\sigma}),\ \ \ \ p_r({\bf \Theta})=\mathcal{N}({\bf \Theta}; {\pmb \mu}^{r}, {\pmb\sigma}^{r})
\end{align}
The expectation log likelihood term in Equation (10) can be efficiently approximated by the Monte Carlo sampling method:
\vspace{-1.2em}

\begin{align}
   \mathcal{L}_{1} \approx \frac{1}{K}\sum_{k=1}^{K}p({\bf \mathcal{D}}|{\bf \Theta}_{k})
\end{align}
\vspace{-1.0em}

\noindent where $K$ is the number of samples and ${\bf \Theta}_{k}$ is the $k$-th sample from distribution $q({\bf \Theta})$.
It has been reported that directly using the mean ${\pmb \mu}$ and variance $\pmb \sigma$ to sample ${\bf \Theta}_{k}$ can make the training process unstable. To address this issue, the reparameterization trick \cite{valet} is adopted to sample ${\bf \Theta}_{k}$ as follows:

\begin{align}
    {\bf \Theta} = {\pmb \mu} + {\pmb \sigma} \odot {\pmb \epsilon}_{k}, \ \ \ \ {\pmb \epsilon}_{k} \sim \mathcal{N({\bf 0}, {\bf I})}.
\end{align}
Under the Gaussian assumption, the second term of Equation (10) can be computed as:
\vspace{-1.0em}

\begin{small}
\begin{align}
    \text{KL}(q({\bf \Theta})||p_{r}({\bf \Theta})) = 
             \sum_{i}\Big{\{}\log\frac{\sigma_{r,i}}{\sigma_{i}} + \frac{\sigma_i^{2}+(\mu_{i}-\mu_{r,i})^{2}}{2\sigma_{r,i}^{2}} - \frac{1}{2}\Big{\}}
\vspace{-1.0em}
\end{align}
\end{small}

\noindent where $\mu_{i}$ and $\sigma_{i}$ are the $i$-th component of ${\pmb \mu}$ and ${\pmb \sigma}$ respectively. 
The gradient of the Bayesian model parameters ${\pmb \mu}$ and ${\pmb \sigma}$ can be computed using the standard back-propagation algorithm as follows:
\vspace{-1.2em}

\begin{align}
    \frac{\partial \mathcal{L}}{\partial \mu_{i}} = \frac{1}{K} \sum_{k=1}^{K} \frac{\partial \mathcal{L}_{1}}{\partial \mu_{i}} - \frac{\mu_{i}-\mu_{r,i}}{\sigma^{2}_{j}} \\ \frac{\partial \mathcal{L}}{\partial \sigma_{i}} =  \frac{1}{K} \sum_{k=1}^{K} \frac{\partial \mathcal{L}_{1}}{\partial \sigma_{i}} - \frac{\sigma^{2}_{i}-\sigma^{2}_{r,i}}{\sigma^{2}_{j}}
\end{align}
\vspace{-2.05em}

\subsection{Implementation details of the Bayesian Transformer LM}
The performance and efficiency of the proposed Bayesian Transformer LMs are affected by the follow details:

\noindent{\bf Position of uncertainty modelling:} 
Although applying Bayesian estimation to all model parameters in the Transformer LM is theoretically feasible, it is practically highly expensive in both model training and evaluation.
To solve this issue, the Bayesian estimation is only applied on part of the model parameters to narrow down the scope of uncertainty modelling. Equation (9) can be re-written as:
\vspace{-1.4em}

\begin{align}
    \log p({\mathcal{D}}) = \log\int p({\bf \mathcal{D}}|{\bf \Theta})p_{r}({\pmb \theta}) d {\pmb \theta}
\end{align}
\vspace{-1.2em}

\noindent where ${\pmb \theta} \in {\bf \Theta}$ is the part of parameters associated with Bayesian estimation. 
We applied Bayesian estimation to the feed forward and multi-head self-attention modules in the Transformer block and the embedding layer respectively.
Specifically, when Bayesian inference is applied on the multi-head attention layers, the query, key and value weight matrices in Equation (2) are assumed to be independent among themselves, thus separate variational distributions are used to model the uncertainty associated with them.

\noindent{\bf Parameter sampling strategy:}
As shown in Equation (12), the Bayesian Transformer LM requires Monte Carlo sampling to approximate the log likelihood. 
The computation cost of the model is linearly increased respect to the number of samples $K$.
To maintain the Bayesian Transformer LM's computation cost comparable to the standard model, we set $K=1$ during the training stage. 
As for evaluation, we use the mean of the Bayesian parameters to approximate Equation (8) as follows: 
\vspace{-1.4em}

\begin{align}
    p({\bf w}_t|{\bf w}_1, ..., &{\bf w}_{t-1}) \approx p({\bf w}_t|{\bf w}_1, ..., {\bf w}_{t-1}, {\bf \Theta}_{mean})
\end{align}
\vspace{-1.2em}

\noindent{\bf Choice of prior distribution:} When training the Bayesian Transformer LM, a suitable choice of the prior is required. In our experiments we used the parameters obtained from a standard Transformer LM as the prior's mean. The prior's variance is set to be ${\bf 1}$. All the Transformer and Bayesian Transformer LMs are interpolated with the $4$gram LM.

\begin{table*}[htb]
\vspace{-1.5em}
    \centering
    \small
    \caption{Perplexity and WER(\%) of the baseline 4-gram (4g) LM, Transformer LM and various Bayesian Transformer LMs before and after further interpolation with the baseline Transformer on NIST Switchboard English eval2000, rt02 and rt03 test sets.  FF, MHA and EMB represent the feed forward, multi-head self-attention and the embedding layer respectively. "+4g" denotes interpolation with 4gram. "$\dag$" denotes statistically significant results were obtained over the Transformer baseline (line 2).}
    \begin{tabular}{c|c|cc|c|cc|ccc|ccc}
    \toprule
     \multirow{2}{*}{ID}&     \multirow{2}{*}{LM}                & \multicolumn{2}{c|}{Bayesian }   & PPL       & \multicolumn{2}{c|}{eval2000} & \multicolumn{3}{c|}{rt02} & \multicolumn{2}{c}{rt03} \\
                        &                    & Block & Position                & (swbd)    & swbd & callhm                & swbd1 & swbd2 & swbd3    & fsh  & swbd               \\
         \hline
         \hline
     1&    4gram                              & \multicolumn{2}{c|}{Not Applied}& -         &9.7   &18.0                   & 11.5  &15.3   & 20.0     & 12.6 & 19.5 \\
     2&    Transformer(+4g)                        & \multicolumn{2}{c|}{Not Applied}& 41.50     &7.9   &15.7                   &  9.5  &12.8   & 17.4     & 10.4 & 17.3 \\
         \hline
         \hline
     3&    \multirow{8}{*}{Bayes Transformer(+4g)} & -      & EMB                    & 41.01     &7.7   &15.6                   &  9.5  &12.6   & 17.1$^\dag$      & 10.2 & 17.1$^\dag$  \\
     4&                                       & 1      & MHA                    & 40.95     &7.7   &15.5                   &  9.5  &\bf12.5$^\dag$    & 17.1$^\dag$      & 10.2 & 17.1$^\dag$  \\
     5&                                       & 1      & FF                     & \bf40.65     &7.7   &\bf15.4$^\dag$                   &  \bf9.4  &12.6$^\dag$    & \bf17.0$^\dag$      &10.2$^\dag$  & \bf17.0$^\dag$  \\
     \cline{3-12}
     6&                                       & 1-2    & FF                     & 41.11     &7.7   &15.6                   &  9.5  &12.6   & 17.2     & 10.3 & 17.1 \\
     7&                                       & 1-3    & FF                     & 42.45     &7.8   &15.8                   &  9.5  & 12.7  & 17.2     & 10.3 & 17.2\\
     8&                                        & 1-4    & FF                     & 47.54     &8.0   &16.0                   &  9.9  & 13.0  & 17.6     & 10.7 & 17.5\\
     9&                                        & 1-5    & FF                     & 54.19     &8.3   &16.2                   &  10.2  & 13.5  & 18.0     & 11.1 & 18.0\\
     10&                                        & 1-6    & FF                     & 74.50     &8.9   &17.3                   &  10.8  & 14.3  & 18.7     & 12.0 & 18.8\\
         \hline
         \hline
     11& \multirow{3}{*}{+Transformer(+4g)}          & -      & EMB                    &     40.03     &7.7   &15.5                   &  9.4  &12.6$^\dag$   & 17.1$^\dag$     & \bf10.1$^\dag$ & 17.0$^\dag$ \\
     12&                                        & 1      & MHA                    &     39.70     &\bf7.6$^\dag$   &15.4$^\dag$                   &  9.3  &\bf12.5$^\dag$    & \bf17.0$^\dag$     & \bf10.1$^\dag$  & \bf16.9$^\dag$  \\
     13&                                        & 1      & FF                     &     \bf39.42     &\bf7.6$^\dag$   &\bf15.2$^\dag$                   &  \bf9.3  &\bf12.5$^\dag$   & \bf17.0$^\dag$     & \bf10.1$^\dag$ & \bf16.9$^\dag$ \\
    \bottomrule
    \end{tabular}
\vspace{-0.8em}
\end{table*}

\section{Experimental setup}
In this section, we present the  details of the datasets used in the experiments before introducing the baseline speech recognition systems. 

\subsection{Datasets}
{\bf Switchboard and Fisher}: The combined Switchboard and Fisher transcriptions adopted in our experiments contain 34M words with a 30k vocabulary lexicon for language modelling.  

\noindent {\bf DementiaBank Pitt}:
The small DementiaBank Pitt transcription \cite{dementia} adopted in our domain adaptation experiments contains 167k. 
A 3.6k words recognition vocabulary was used. 
\vspace{-0.6em}
\subsection{Baseline Transformer LMs}
The standard and Bayesian Transfomer LMs used in our experiments contain 6 Transformer blocks with 4096 hidden nodes in the feed forward module and 512 dimension for the residual connection.
The output dimensionality of the word embedding layer is set to be 512 and the input dimensionality is set to be equal to the vocabulary size of the dataset.
Pytorch was used to implement the Transformer LMs. 
The model parameters are optimized using stochastic gradient descent (SGD) optimizer with initial learning rate 0.1. 
All Transformer LMs in our experiments are on word level.
We use 1 Nvidia V100 GPUs to train the LMs.

\vspace{-0.6em}
\subsection{Baseline Speech Recognition Systems}
{\bf Switchboard system:} Following the Kaldi \cite{kaldi} recipe\footnote{Kaldi: egs/swbd/s5c/local/chain/tuning/run tdnn 7q.sh}, in the Switchboard experiments, the speech recognition system used to generate the N-best list for rescoring was based on factorized time-delay neural networks (TDNN-Fs) \cite{tdnnf} featuring speech perturbation, i-Vector, LHUC speaker adaptation and Lattice-free maximum mutual information (LF-MMI) \cite{lfmmi} sequence training.  

\noindent {\bf DementiaBank Pitt system:} The speech recognition system used the DementiaBank Pitt experiments was similar to the TDNN-F based Switchboard system with additional domain adaptation. Details of this system can be found in \cite{dementia}.

\section{Experiments}
In this section, we present our experimental results in terms of perplexity (PPL) and word error rate (WER) for the proposed Bayesian Transformer LMs on the Switchboard and DementiaBank corpora.
\vspace{-0.6em}

\subsection{Experiments on the Switchboard Corpus}
\label{EXP:1}
Table 1 presents the experimental results of the proposed Transformer language model on the Switchboard corpus.
Several trends can be observed from Table 1: 
1) The proposed Bayesian Transformer LMs (line 3-5) outperform the baseline Transformer language model (line 2) in terms of both perplexity and word error rate. 
2) Applying the Bayesian estimation on the feed forward (FF) module outperforms using Bayesian estimation on the multi-head self-attention (MAH)  module and the embedding (EMB) layer in terms of the PPL and WER; 
3) Compared with applying Bayesian estimation to multiple Transformer blocks (line 6 - 10), adopting Bayesian estimation only on the lowest Transformer block (line 5) produced the best PPL and WER performance. 
One possible explanation of this observation is that the parameters associated with the higher Transformer blocks are expected to be more deterministic than those in the lower blocks, while the larger part of the underlying data variability is expected at the lowest block immediately after the embedding layer.
4) Further performance improvements can be obtained by interpolating the Bayesian Transform LM with the standard Transformer LM (line 11-14). 

\noindent The Bayesian Transformer LM with parameter uncertainty modelled at the lowest feed forward layer produced the best performance after interpolation with both the 4-gram and baseline Transformer (line 13). Statistically significant WER reductions of 0.3-0.5\% were obtained across all the data sets except the {\tt swbd1} portion of {\tt rt02} over the baseline Transformer LM (line 2). The statistical significance test was conducted at level  0.5 based on matched pairs sentence-segment word error (MAPSSWE) for recognition performance analysis.

To further analyse the proposed Bayesian Transformer LM’s ability in reducing the risk of overfitting and improving generalization, Figure 2 compares the performance between the proposed Bayesian Transformer LM (line 5 in Table 1) and the standard Transformer with and without the dropout operation.
As shown in Figure 2, the proposed Bayesian LM consistently outperforms the other two LMs with varying feed forward module dimensionality from 512 to 16384. 


\begin{figure}[htb]
\vspace{-1.0em}
\newcommand{\tabincell}[2]{\begin{tabular}{@{}#1@{}}#2\end{tabular}}
\setlength{\parskip}{0.0ex}
  \centering
  \centerline{\includegraphics[width=6.5cm]{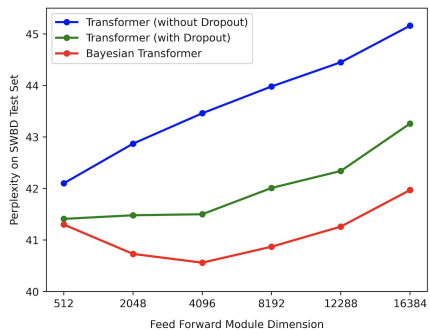}}
\caption{Perplexity on SWBD test data obtained using the baseline and the Bayesian Transformer LMs with varying feed forward layer dimensionality.}
\label{fig:res}
\vspace{-1.5em}
\end{figure}

\vspace{-1.0em}
\subsection{Experiments on the DementiaBank Pitt Corpus}
\label{EXP:3}
The PPL and WER results on the DementiaBank Corpus are presented in Table 2.
The $4$gram, Transformer and Bayesian Transformer LMs were first trained using the combined 2.4M words DementiaBank Pitt, Switchboard and Fisher transcriptions. 
To reduce the domain mismatch between the three corpora, the baseline and Bayesian Transformer LMs were further adapted to the Pitt transcripts only by using either fine-tuning or Bayesian adaptation.
This two adapted Transformer LMs are shown in line 3 and 5 in Table 2 respectively.
The fine-tuning adapted model parameters in line 3 served as the prior of the Bayesian Transformer adaptation in line 5.
It can be observed from Table 2 that the Bayesian adapted Transformer LM outperforms the fine-tuned Transformer LM by 0.37\% absolute WER reduction. 

\begin{table}[htb]
\vspace{-0.5em}
    \centering
    \caption{PPL and WER(\%) results on the DemntiaBank Pitt Corpus. "finetune" means fine-tuning the model parameters using only the DemntiaBank Pitt Corpus. "bayes-adapt" mean adapt the Bayesian Transformer using only the DemntiaBank Pitt Corpus LM with the parameters in system 4 as prior.   "+4g" denotes interpolation with 4gram. "$\dag$" denotes statistically significant results were obtained over the system 3.}
    \begin{tabular}{c|c|c|c|cccc}
    \toprule
      ID&   LMs     & Adapt & PPL  & WER(\%)\\
         \hline
         \hline
     1&    4gram    &  \xmark &17.07 & 30.67\\
        \hline
     2&   \multirow{2}{*}{Transformer(+4g)} & \xmark & 21.83 & 30.65 \\
     3&                                & fine-tuning & 14.56 & 30.25\\
        \hline
        \hline
     4&   \multirow{2}{*}{Bayes Transformer(+4g)} & \xmark &19.88 & 30.49\\
     5&                                      & bayes-adapt & 13.99 & \bf29.88$^\dag$\\
    \bottomrule
    \end{tabular}
    \label{tab:my_label}
\vspace{-0.8em}
\end{table}
\vspace{-1.0em}

\section{Conclusion}
\label{Con}

This paper presents a Bayesian learning framework for Transformer language model estimation to improve their generalization performance.
Consistent performance improvements in terms of both perplexity and WER were obtained on the Swithboard and DementiaBank Pitt datasets, thus demonstrating the advantages of the proposed Bayesian Transformer LMs for speech recognition.

\section{Acknowledgement}
\label{Acknowledgement}

This research is supported by Hong Kong RGC GRF grant No. 14200218, 14200220, Theme-based Research Scheme T45-407/19N, Innovation \& Technology Fund grant No. ITS/254/19, and Shun Hing Institute of Advanced Engineering grant No. MMT-p1-19.

\vfill\pagebreak

\clearpage
\bibliographystyle{IEEEbib}
\small
\bibliography{reference}
\end{document}